# On the Complexity of Policy Iteration

Yishay Mansour* and Satinder Singh
AT&T Labs-Research
180 Park Avenue
Florham Park, NJ 07932-0971
{mansour,baveja}@research.att.com

## Abstract

Decision-making problems in uncertain or stochastic domains are often formulated as Markov decision processes (MDPs). Policy iteration (PI) is a popular algorithm for searching over policy-space, the size of which is exponential in the number of states. We are interested in bounds on the complexity of PI that do not depend on the value of the discount factor. In this paper we prove the first such non-trivial, worst-case, upper bounds on the number of iterations required by PI to converge to the optimal policy. Our analysis also sheds new light on the manner in which PI progresses through the space of policies.

## 1 Introduction

The problem of decision-making in uncertain or stochastic environments is central to artificial intelligence (AI) [7, 6]. The framework of Markov decision processes (MDPs) developed in the operations research community [1] is increasingly used within AI to formulate such problems. In this formulation, the environment is assumed to be in one of a finite-set of states, the decision-making agent has a choice of actions in each state of the environment, executing an action causes a stochastic change in the state of the environment, and the agent receives a stochastic reward in return for executing the action. The agent's goal is to choose actions so as to maximize a cumulative discounted measure of rewards over some time horizon. Here we consider the planning problem in which we are given a full description of the MDP and have to compute the optimal action-selection policy.

One reason for the popularity of the MDP framework within AI is the availability of a number of well-studied classes of algorithms for planning in MDPs: linear-programming [2], value iteration [1], and policy iteration [3]. Linear programming and value iteration are known to compute the optimal policy in time polynomial in the size of the representation of the MDP and the discount factor [4, 2]. While no direct analysis of policy iteration is available, one can bound the number of steps of "greedy" policy iteration (which greedily accepts all single-state action changes that are improvements) by the number of steps of value iteration. This implies that policy iteration also runs in time polynomial in the size of the representation and the discount factor [3, 2].

However, our goal is to derive bounds for solving MDPs that do not depend on the discount factor. For value iteration the dependence on the discount factor is unavoidable. For linear programming, in general, it is a major open problem whether there exists a strongly polynomial algorithm, i.e., runs in time polynomial in the number of parameters and independent of the size of the representation of the parameters. For PI we can bound the number of steps independent from the representation size and discount factor as follows: PI is guaranteed to improve the policy at every step and therefore the total number of steps is trivially upper-bounded by the total number of policies. This bound is of course independent of the discount factor. However, note that the total number of policies is exponential in the number of states.

In this paper, we prove the first non-trivial upper bound on the worst-case number of steps PI can take. For the specific case of $n$ states and two actions the total number of policies is exactly $2^n$. We show that "greedy" PI will take at most $O(\frac{2^n}{n})$ steps. We also define a randomized PI (which accepts each single-state action change that is an improvement with probability 0.5) and prove that in the worst-case it will take at most $O(2^{0.78n})$ steps. For the general case of $k$ actions we show an bound of $O(k^n/n)$ for greedy PI and $O([(1 + \epsilon_k)k/2]^n)$ for random PI (where $\epsilon_k$ is small for large $k$ and will be defined later). Note that these bounds are independent of the size of representation of the specific parameters of the MDP and in particular do not depend on the discount factor. Our analysis also sheds new light on the manner in which PI progresses through the space of policies.

---
*On sabbatical from Tel-Aviv University.

We view our results as the first step towards a better understanding of PI. This is an important issue because there is strong empirical evidence in favor of PI over value iteration and linear programming in solving MDPs [4]. While in practice it is difficult to construct MDPs for which greedy PI takes more than $n$ steps, no general rigorous lower bounds are known. A lower-bound is known for a particular form of PI, called sequential PI (which at each step accepts only one of the single-state action changes that are improvements) — in the worst-case, sequential PI can take $\Omega(2^n)$ steps on a two-action MDP, when the adversary controls which improvements are selected [5, 4]. However it is not clear whether results about sequential PI transfer to other forms of PI, e.g., greedy or random PI. In fact our results show that there is a gap between the worst-case complexities of sequential and both greedy and random PI.

This paper is organized as follows: Section 2 defines the MDP model and its notation. Section 3 introduces the general scheme of policy iteration and proves a few general results concerning it. Section 4 derives an upper-bound on the time complexity of greedy PI, while Section 5 derives an upper-bound to the time complexity of random PI (both for two-action MDPs). Section 6 extends our results to a general multi-action MDP. Section 7 concludes with a summary of our contributions and open problems.

## 2 Model

In this section we define the Markov decision process (MDP) framework.

**Definition 1** *An MDP is a tuple $(S, A, P, R)$: $S$ is a finite set of states the environment can be in, $A$ is a finite set of actions available to the agent, $P$ is the table of transition probabilities, where $P(s'|s, a)$ is the probability of a transition to state $s'$ upon executing action $a$ in state $s$, and $R$ is the reward function, where $R(s, a)$ is the expected reward received by the agent upon executing action $a$ in state $s$.*

We define the agent's *return* to be the discounted sum of rewards over an infinite horizon, i.e., we use the infinite-horizon discounted framework in this paper. More formally, the agent's return is $\sum_{t=0}^{\infty} \gamma^t r_t$ where $r_t$ is reward received at time step $t$, and $0 \leq \gamma < 1$ is a discount factor that makes future reward less valuable than immediate reward. The agent's goal is to select actions so as to maximize its expected return. In infinite-horizon discounted MDPs the agents expected return is maximized by a policy (a mapping from states to actions), called the optimal policy.

Useful quantities in analyzing MDP-decision-making are value functions: one defined over states and the other defined over state-action pairs.

**Definition 2** *Let $V^\pi(s)$ be the expected return if the start-state is $s$ and the agent executes policy $\pi$ forever. Let $Q^\pi(s, a)$ be the expected return if the start-state is $s$ and the agent executes action $a$ to begin with and thereafter follows policy $\pi$.*

Note that by the definition above $Q^\pi(s, a) = R(s, a) + \gamma \sum_{s'} P(s'|s, a) V^\pi(s')$. The agent's goal, restated in terms of value functions, is that of finding an optimal policy $\pi^*$ that satisfies $\pi^* = argmax_\pi V^\pi$. The optimal value function $V^{\pi^*}$ is denoted simply as $V^*$ and the associated Q-value function as $Q^*$. Note that there can be more than one optimal policy, however, $V^*$ and $Q^*$ are unique.

The total number of policies in an MDP is $k^n$, where $n = |S|$ and $k = |A|$. In most of the paper we discuss the case that there are only two actions, i.e. $|A| = 2$, which implies that the number of policies is bounded by $2^n$. At the end of the paper we discuss the general case, where $|A| = k$.

## 3 General Policy Iteration

General policy iteration works as follows. At each iteration consider changing the action at each state while keeping the actions for all the other states fixed to the current policy. Some such single-state action changes will improve upon the current policy. Different variants of policy iteration differ in which single-state improvements they accept at each step.

Before we can describe general PI, we must define what it means for one policy to be better than another. We define a partial order between the policies as follows.

**Definition 3** *For two policies, $\pi$ and $\pi'$, we have $\pi \succ \pi'$ if for each state $s$, $V^\pi(s) \geq V^{\pi'}(s)$, and for some state $s$, $V^\pi(s) > V^{\pi'}(s)$. If for every state $s$ we have $V^\pi(s) = V^{\pi'}(s)$ then $\pi \approx \pi'$.*

The partial ordering tells us when a policy is better than another and when they are incomparable. Clearly any optimal policy is better than all suboptimal policies and equivalent ($\approx$) to all other optimal policies. This partial order is central to our analysis.

Given a policy, $\pi$, we can define, using the function $Q^\pi$, the single-state improvements that could improve that policy. The following definition gives the necessary notation that we use later.

**Definition 4** *Given a policy $\pi$, let the modification set $T^\pi \subset S \times A$ be the set of all pairs $(s,a)$ such that changing the action of $\pi$ to $a$ in state $s$ improves the return of the policy, i.e. $Q^\pi(s,a) > V^\pi(s)$. We define $\text{states}(T^\pi)$ to be the states that appear in $T^\pi$, i.e., $\{s : (s,a) \in T^\pi\}$. If each state appears only once in $T^\pi$ we say that $T^\pi$ is well defined.*

Let $\pi$ be a policy such that $T^\pi$ is well defined. (Note that if the MDP has only two action then for any policy $\pi$ we have that $T^\pi$ is well defined.) For a set $U \subset T^\pi$ let $\text{modify}(\pi, U)$ define a policy $\pi'$ whose actions are the same as those of policy $\pi$ on states not in $\text{states}(U)$ and $\pi'(s) = a$ for $(s,a) \in U$.

Figure 1 presents the general *policy iteration* algorithm. In every iteration there are two basic steps: the first, Improvement Selection Step, selects which single-state improvements to make, and the second, Policy Improvement Step, modifies the policy accordingly. Different methods for selecting subsets of $T^\pi$ to modify the policy lead to different PI algorithms.

The following two theorems are well known properties of general policy iteration. The first claims that accepting any non-zero number of single-state improvements can only improve the policy, and the second claims that there always exists at least one single-state improvement that improves the policy, unless the policy is already optimal. (For proofs see, e.g., [2].)

**Theorem 1** *For any $U \subset T^\pi$, let $\pi' = \text{modify}(\pi, U)$. If $U \neq \emptyset$ then $\pi' \succ \pi$.*

**Theorem 2** *For any sub-optimal $\pi$, $T^\pi \neq \emptyset$.*

The above two theorems immediately imply that all instantiations of general PI strictly improve the policy at every iteration. Therefore, every iteration from a current to a next policy at least skips, or rules out, all the policies that are equal to, or better than, the current policy and worse than the next policy. How many such policies are there at every iteration? There is at least one such policy: the current policy itself. This, of course, implies an upper bound of $k^n$ steps. For specific improvement-selection methods, defined in the following sections, we perform a more careful analysis of the number of equal or better policies that get ruled out at each iteration. The more policies we can rule out at each iteration the better the upper-bound will be. Our analysis will be mainly based on properties derived from Theorems 1 and 2.

In the rest of this section, we prove a few properties that hold for all instances of PI. The first is actually a property of the partial order itself: in general two policies may be incomparable but if they differ only in one state then they must be comparable.

**Lemma 3** *Let $\pi$ and $\pi'$ be two policies whose actions differ in only one state $s$, i.e., $\pi(u) = \pi'(u)$ for $u \neq s$. Then either $\pi \succ \pi'$, $\pi' \succ \pi$, or $\pi \approx \pi'$.*

**Proof:** If $Q^\pi(s, \pi'(s)) > V^\pi(s)$ then $\pi' \succ \pi$. If $Q^\pi(s, \pi'(s)) < V^\pi(s)$ then $\pi \succ \pi'$. Otherwise $\pi \approx \pi'$. $\square$

The following lemma gives an interesting connection between the optimality of a policy $\pi$ and the states in its modification set $T^\pi$.

**Lemma 4** *For any policy $\pi$, and any policy $\pi'$ that is identical to $\pi$ on states in $\text{states}(T^\pi)$, either $\pi \succ \pi'$, or $\pi \approx \pi'$.*

**Proof:** Consider an MDP $M'$ such that the only action possible from a state $s \in \text{states}(T^\pi)$ is $\pi(s)$. Clearly both $\pi$ and $\pi'$ are valid policies for $M'$. On the other hand in $M'$ there is no local improvement for $\pi$, i.e. $T^\pi_{M'} = \emptyset$. By Theorem 2, $\pi$ is optimal for $M'$. Therefore $\pi \succ \pi'$ (or $\pi \approx \pi'$). $\square$

For an MDP with two actions we can show that PI,

> Set $\pi_0$ to an arbitrary policy.
> **WHILE** $T^{\pi_i} \neq \emptyset$ **DO**
>     Improvement Selection Step: $U \leftarrow \text{select}(T^{\pi_i})$
>     Policy Improvement Step:    $\pi_{i+1} \leftarrow \text{modify}(\pi_i, U)$.
>     $i \leftarrow i + 1$.
> **OUTPUT** $\pi_i$.

Figure 1: General Policy Iteration Algorithm. The only assumption we make about the function `select` is that it returns a non-empty subset of its argument and at most one action for every state.

at different iterations, considers an improvement over a different subset of the states. This result is general to any PI.

**Lemma 5** *During a run of general policy iteration algorithm on a two-action MDP, there are no $i$ and $j$, $i < j$, such that $\text{states}(T^{\pi_i}) \subseteq \text{states}(T^{\pi_j})$.*

**Proof:** We prove the lemma by contradiction. Assume that there exists $i$ and $j$, $i < j$ such that $\text{states}(T^{\pi_i}) \subseteq \text{states}(T^{\pi_j})$. Let $T = \text{states}(T^{\pi_i}) \subseteq \text{states}(T^{\pi_j})$. Let $U' = \{(s, a) : a = \pi_i(s) \text{ and } \pi_i(s) \neq \pi_j(s) \text{ and } s \in T\}$. Clearly $U' \subseteq T^{\pi_j}$, since there are only two actions. Then $\pi' = \text{modify}(\pi_j, U')$ is identical with $\pi_i$ on the states in $T = \text{states}(T^{\pi_i})$. Therefore, by Lemma 4 we have that $\pi_i \succ \pi'$ or $\pi_i \approx \pi'$. This contradicts the fact that $\pi' \succ \pi_j \succ \pi_i$. □

So far we have showed that a subset of states can appear at most once in general policy iteration, when the MDP has only two actions. This still leaves open the possibility that all subsets appear in the run of the algorithm, and thus we observe all $2^n$ policies, for a two-action MDP. The next step is to show that each time we perform modify on a large subset of the states we rule out many policies.

## 4 Greedy Policy Iteration

Greedy policy iteration is PI with $\text{select}(T) = T$, namely, we perform all the possible single-state action improvements at each policy improvement step. (We assume that $T$ is well defined, which is always the case for two-action MDPs. For the general case see Section 6.)

The next lemma shows that each time we perform a modify operation we rule out a number of policies that is at least the the size of the modification set.

**Lemma 6** *Let $\pi$ be a policy such that $T^\pi$ is well defined, and $\pi' = \text{modify}(\pi, T^\pi)$. Then there are at least $|T^\pi|$ policies $\pi_i$, $1 \leq i \leq |T^\pi|$, such that $\pi' \succeq \pi_i \succ \pi$.*

**Proof:** We show by induction on $m$, that if $|T^\pi| \geq m$ then there are at least $m$ policies $\pi_i$, $1 \leq i \leq m$, such that $\pi' \succeq \pi_i \succ \pi$. The base of the induction, $m = 1$, follows from Theorem 1.

For the inductive step we assume that the claim holds for $m-1$ and we show that it holds for $m$. Assume that $|T^\pi| \geq m$. Consider all the single state modifications to $\pi$ using $T^\pi$, i.e., consider all $Z_j$, such that $Z_j \subset T^\pi$ and $|Z_j| = 1$. Let $U_1 = Z_j$ such that for any $Z_i$, we have that $\text{modify}(\pi, Z_j) \not\succ \text{modify}(\pi, Z_i)$. (Note that $Z_j$ is not necessarily unique, since we have a partial order.) Let $\pi_1 = \text{modify}(\pi, U_1)$.

With out loss of generality, let $U_1 = \{(s_1, b_1)\}$. For any other pair $(s_i, b_i) \in T^\pi$, for $i > 1$, we show that $(s_i, b_i) \in T^{\pi_1}$. Let $\pi_1^i = \text{modify}(\pi_1, \{(s_i, b_i)\})$. By Lemma 3 we know that either $\pi_1 \succ \pi_1^i$ or $\pi_1^i \succeq \pi_1$. We would like to claim that the relation $\pi_1 \succ \pi_1^i$ is not possible.

For contradiction assume that $\pi_1 \succ \pi_1^i$. Consider $q_i = \text{modify}(\pi, \{(s_i, b_i)\})$. Note that $\pi_1^i = \text{modify}(\pi, \{(s_1, b_1), (s_i, b_i)\})$, and therefore by Lemma 3 we know that either $q_i \succ \pi_1^i$ or $\pi_1^i \succeq q_i$. If $\pi_1^i \succeq q_i$ then $\pi_1 \succ \pi_1^i \succeq q_i$ contradicting the minimality of $\pi_1$. Therefore $q_i \succ \pi_1^i$. Let $\pi(s_1) = a_1$ and $\pi(s_i) = a_i$. Since $\pi_1 \succ \pi_1^i$, this implies that $(s_i, a_i) \in T^{\pi_1^i}$ and similarly, since $q_i \succ \pi_1^i$, this implies that $(s_1, a_1) \in T^{\pi_1^i}$. By Theorem 1 this implies that

$$\pi = \text{modify}(\pi_1^i, \{(s_1, a_1), (s_i, a_i)\}) \succ \pi_1^i,$$

contradicting the fact that $\pi_1^i \succ \pi$. Hence, $\pi_1^i \succ \pi_1$.
This implies that $(s_i, b_i) \in T^{\pi_1}$, for $i > 1$. Therefore, we have $|T^{\pi_1}| \geq |T^\pi| - 1 \geq m - 1$. The lemma follows from the inductive hypothesis on $\pi_1$. □

We can now state and prove our upper-bound on the number of steps of greedy policy iteration.

**Theorem 7** *The greedy policy iteration algorithm considers at most $O(2^n/n)$ different policies for a 2-action MDP.*

**Proof:** The analysis has two parts. The first part includes the case where the set $T^\pi$ is small. For this case we simply show that there are very few such policies. The second case will include the cases when $T^\pi$ is large. For this case we show that each iteration eliminates $\Omega(n)$ policies, that have not been eliminated before.

We define a set to be small if $|T^\pi| \leq n/3$. By Lemma 5 we do not consider the same set of states twice. This bounds the number of such modifications by

$$\sum_{k=0}^{n/3} \binom{n}{k} \leq 2\binom{n}{n/3} \leq 3\frac{2^n}{n},$$

where the second inequality holds for $n \geq 3$. (The first inequality follows from the fact that for $k < n/3$ we have that $\binom{n}{k}/\binom{n}{k-1} > 2$.)

For policies $\pi_i$ such that $|T^{\pi_i}| \geq n/3$, by Lemma 6 we have that at least $n/3$ policies better than or equal to our current policy are ruled out after this iteration. This implies that the total number of policies that we consider is bounded by,

$$3\frac{2^n}{n} + \frac{2^n}{n/3} = 6\frac{2^n}{n},$$

where the first term is the number of policies with small number of improvements and the second term is a bound on the number of policies with a large number of improvements. □

## 5 Random Policy Iteration

Formally, random policy iteration defines $\texttt{select}(T)$ as a random subset of $T$ where each subset has probability $2^{-|T|}$. (We assume that $T$ is well defined. For the general case seee Section 6.) Intuitively, we can think of random policy iteration as deciding to accept each local improvement with probability half. Even though we allow for the empty subset for convenient proofs, in practice one may ignore such iterations.

The property that we would like to prove is that for a two-action MDP the number of policies that we rule out after considering each policy is at least $2^{|T^{\pi_i}|-1}$ rather than only $|T^{\pi_i}|$ (as in greedy policy iteration). This enables us to improve our bound on the running time significantly.

We first show another property of general PI: that no policy $\pi'$ incomparable to $\pi_i$ is ever considered after iteration $i$.

**Lemma 8** *Consider a run of a general policy iteration algorithm, and let $\pi_i$ be the policy at iteration $i$. Let $\pi'$ be a policy such that $\pi' \not\succ \pi_i$. For any $j > i$ we have that $\pi_j \neq \pi'$.*

**Proof:** By Theorem 1 we know that for each $j$ we have that $\pi_j \succ \pi_{j-1}$. By transitivity, we have that $\pi_j \succ \pi_{i+1}$, for $j > i + 1$. Since $\pi' \not\succ \pi_i$, it implies that $\pi' \neq \pi_j$. □

From the above lemma we know that the only policies that we can reach after $\pi_i$ are policies that are comparable with $\pi_i$. This implies that any policy which is either strictly inferior to $\pi_i$, or incomparable to $\pi_i$ will never be considered. The next step is to argue that the number of policies that we rule out at phase $i$ has an expected value of at least $2^{|T^{\pi_i}|-1}$. We first prove a general property of selecting a random element in a partial order.

**Lemma 9** *Let $\succ$ be a partial order over $\Pi$. If we chose a random element $r \in \Pi$, with uniform probability, then the expected number of elements $s \in \Pi$ such that $s \succ r$ is at most $|\Pi|/2$.*

**Proof:** For any element $v \in \Pi$ we associate two sets. $\Pi_v^+$ includes all the elements $s$ such that $s \succ v$, and $\Pi_v^-$ includes all the elements $s$ such that $v \succ s$. For every pair of elements $v_1 \succ v_2$ we have that $v_1 \in \Pi_{v_2}^+$ and $v_2 \in \Pi_{v_1}^-$. This implies that

$$\sum_{v \in \Pi} |\Pi_v^+| = \sum_{v \in \Pi} |\Pi_v^-| \leq \frac{|\Pi|^2}{2}.$$

Therefore the expected value of $|\Pi_r^+|$ is at most $|\Pi|/2$. □

The following corollary combines Lemma 8 and Lemma 9.

**Corollary 10** *Consider a run of the random policy iteration algorithm on a two-action MDP. Let $\pi_i$ be the policy at iteration $i$, then the expected number of policies $\pi'$, such that $\pi_{i+1} \succ \pi' \succ \pi_i$ is at least $2^{|T^{\pi_i}|-1}$.*

Unfortunately it is not true that at each step we expect to rule out $\Omega(2^{|T^{\pi_i}|})$ policies, with high probability. Rather we can say that there is some constant probability that this will happen, and then claim that in a run with $m$ iterations we should have, with high probability, this occurring $\Omega(m)$ times.

**Theorem 11** *The random policy iteration algorithm, for a two-action MDP, considers at most $O(2^{0.78n})$ different policies, with probability $1 - 2^{-2^{\Omega(n)}}$.*

**Proof:** As before we consider two cases, that of small sets and that of large sets. We define a set to be small if $|T^{\pi_i}| \leq pn$, where the constant $p > 0$ will be selected later. As before we bound the number of iterations with small sets by $\sum_{i=0}^{pn} \binom{n}{i} \leq 2^{H(p)n+1}$, where $H(p)$ is the binary entropy, i.e. $H(p) = -p \log p - (1-p) \log(1-p)$.

Now we are interested in bounding the number of iterations with large sets. Assume that we have $m$ such iterations. By Corollary 10 the expected number of policies we rule out is at least $2^{pn-1}$ policies, at each such iteration. This implies that with probability $1/3$ we rule out at least $2^{pn-2}$ policies. (If this occurs with probability strictly less than $1/3$, then the expected number of policies we rule out is strictly less than $(1/3)2^{pn} + (2/3)2^{pn-2} = 2^{pn-1}$, which contradicts Corollary 10.)

An iteration with a large set is *good* if it chooses a set that rules out at least $2^{pn-2}$ policies. From above, the probability that an iteration is good is at least $1/3$. A run is called typical if at least $m/4$ of the $m$ iterations with large sets are good. The number of large set iterations in a typical run is bounded by $2^{(1-p)n+4}$. The total number of iterations in a typical run is bounded by,

$$2^{H(p)n+1} + 2^{(1-p)n+4} \leq 2^{0.78n},$$

for $p = 0.227$ and sufficiently large $n$.

The probability that a run is not typical is at most $e^{-(1/3-1/4)^2 m}$. We are interested in runs in which $m \geq 2^{(1-p)n+4}$, in which case the probability is bounded by $2^{-2^{\Omega(n)}}$. □

## 6 Multi-Action MDPs

In this section we extend the results from two-action to $k$ actions, where $k \geq 2$. Recall that since we have $k$ actions the total number of policies is $k^n$.

First we observe that when there are more than two actions, it might be the case that we have in $T^\pi$ a number of different pairs with the same state, i.e. $T^\pi$ is not well defined. We assume that $T^\pi$ is reduced to $L^\pi$, such that each state appears only in one pair, i.e. $L^\pi$ is well defined. Formally, $L^\pi \subset T^\pi$ and $\text{states}(L^\pi) = \text{states}(T^\pi)$. We do not make any other assumption on the way $L^\pi$ is chosen, and assume that the various PI algorithms perform $U \leftarrow \text{select}(L^\pi)$.

Using the Lemma 4 we can derive the following lemma.

**Lemma 12** *During a run of a general policy iteration algorithm, there are no $i$ and $j$, $i < j$, such that $\text{states}(T^{\pi_i}) \subseteq \text{states}(T^{\pi_j})$ and for every $s \in \text{states}(T^{\pi_i})$ we have $\pi_i(s) = \pi_j(s)$.*

**Proof:** The proof is by contradiction. Assume that such $i$ and $j$ exists. By Lemma 4 we have that $\pi_i \succ \pi_j$ or $\pi_i \approx \pi_j$. By Theorem 1 we have that $\pi_j \succ \pi_i$, since $j > i$, and therefore we have a contradiction. □

The above lemma is the main difference between the two-action case and the multi-action case. This difference results in slightly worse bounds. As in the two action case, our analysis separates the modification sets to small and large. The following corollary of Lemma 12 is used to bound the number of small modifications.

**Corollary 13** *During a run of a general policy iteration algorithm, the number of iterations in which $|L^{\pi_i}| \leq d$ is bounded by $\sum_{j=0}^{d} \binom{n}{j} k^j$.*

We start by bounding the number of iterations, in the worst case, performed by the greedy policy iteration algorithm. Note that Lemma 6 applies to $L^\pi$, since $L^\pi$ is well defined. The following theorem bounds the number of iterations for the greedy policy iteration algorithm in the multi-action case. (The proof is in the same spirit as the two-action case, but the constants are different.)

**Theorem 14** *The greedy policy iteration algorithm considers at most $O(k^n/n)$ different policies.*

**Proof:** As in the proof of Theorem 7, the analysis has two parts. The first part includes the case where the set $L^\pi$ is small. For this case we simply show that there are very few such policies. The second case includes the case when $L^\pi$ is large. For this case we show that each iteration eliminates $\Omega(n)$ policies, that have not been eliminated before.

We define a modification set to be small if $|L^\pi| \leq pn$, where $p = 1/10$. By Corollary 13, the number of small modification sets is bounded by,

$$\sum_{j=0}^{pn} \binom{n}{j} k^j \leq 2\binom{n}{pn} k^{pn} \leq 3\frac{k^n}{n},$$

for $k \geq 2$ and $n \geq 1$.

For policies $\pi_i$ such that $|L^{\pi_i}| \geq pn$, by Lemma 6 we have that at least $pn$ policies better than or equal to our current policy are ruled out after this iteration. This implies that the maximum number of policies that greedy PI considers is bounded by,

$$3\frac{k^n}{n} + \frac{k^n}{pn} = 13\frac{k^n}{n},$$

where the first term is the number of policies with small number of improvements and the second term is a bound on the number of policies with a large number of improvements. □

We now show the bound for random policy iteration. First note that Lemma 8 holds for $L^\pi$, since $L^\pi$ is well defined. In addition Lemma 9 is a general property of partial orders. Therefore, we can derive a corollary similar to Corollary 10.

**Corollary 15** *Let $\pi_i$ be the policy at iteration $i$, then the expected number of policies $\pi'$, such that $\pi_{i+1} \succ \pi' \succ \pi_i$ is at least $2^{|L^{\pi_i}|-1}$.*

Now we can derive the theorem for the random policy iteration algorithm for the multi-action case.

**Theorem 16** *The random policy iteration algorithm considers at most*

$$O\left(\left((1 + \frac{2}{\log k})\frac{k}{2}\right)^n\right)$$

*different policies, with probability $1 - 2^{-\Omega((k/2)^n)}$.*

**Proof:** As in Theorem 11 we consider two cases, that of small sets and that of large sets. We define a set to be small if $|L^{\pi_i}| \leq pn$, where $p = 1 - 2/\log k$. By Corollary 13, the number of iterations with small sets is bounded by $\sum_{i=0}^{pn} \binom{n}{i} k^i \leq 2^n k^{pn}$.

Now we are interested in bounding the number of iterations with large sets. Assume that we have $m$ such iterations. By Corollary 10 the expected number of policies we rule out is at least $2^{pn-1}$ policies, at each such iteration. This implies that with probability $1/3$ we rule out at least $2^{pn-2}$ policies.

An iteration with a large modification set is *good* if it chooses a set that rules out at least $2^{pn-2}$ policies. From above, the probability that an iteration is good is at least $1/3$. A run is called typical if at least $m/4$ of the $m$ iterations with large sets are good. The number of large set iterations in a typical run is bounded by $k^n/2^{pn-4}$. The total number of iterations in a typical run is bounded by,

$$2^n k^{pn} + \frac{k^n}{2^{pn-4}} \leq (\frac{k}{2})^n + 16\left(\frac{k}{2}2^{2/\log k}\right)^n$$
$$\leq 17\left(\frac{k}{2}(1 + \frac{2}{\log k})\right)^n$$

for $k \geq 2$ and $n \geq 1$.

The probability that a run is not typical is at most $e^{-(1/3-1/4)^2 m}$. We are interested in runs in which $m \geq (k/2)^n$, in which case the probability is bounded by $2^{-\Omega((k/2)^n)}$. □

## 7 Conclusion

In this paper we developed a proof technique for deriving upper-bounds on the number of steps required by policy iteration to find an optimal policy. Using our proof technique we are able to establish non-trivial upper-bounds for two important variations of policy iterations.

For greedy policy iteration we proved an upper-bound of $O(\frac{2^n}{n})$, and for random policy iteration we proved an upper-bound of $O(2^{0.78n})$, both in the case that the MDP has two actions. This should be contrasted with the lower-bound of $\Omega(2^n)$ for sequential policy iteration [5, 4]. For the case of $k$ actions we give upper-bounds of $O(\frac{k^n}{n})$ and $O([(1+\epsilon_k)k/2]^n)$, for the greedy and random policy iteration algorithms, respectively.

We have no reason to believe that our bounds are tight. One case where our bounds seems to be "losing" considerably is the following. When counting policies that we rule out we consider only policies that we can reach from $\pi_i$ using its modification set $T^{\pi_i}$. However, in many cases we can rule out additional policies. Another constraint that we were not able to utilize is the benefit of having small modification sets. For example if $T^{\pi_i} = \{(s,a)\}$ then in the two-action case we can rule out half of the possible modification sets. More precisely, we will never have to update the action of state $s$ again. Unfortunately, we did not find a way to take advantage of this property, and we use Lemma 5 only in the sense that the modification sets cannot be equal, rather than the subset property.

It would have been of great benefit if we had good lower bounds for general policy iteration, but unfortunately we do not know of any bound other than the trivial lower-bound of $n$. The gap between upper and lower bounds is still very large and is an interesting subject for future research.

# References

[1] D. P. Bertsekas. *Dynamic Programming: Deterministic and Stochastic Models*. Prentice-Hall, Englewood Cliffs, NJ, 1987.

[2] D. P. Bertsekas. *Dynamic Programming and Optimal Control*. Athena Scientific, Belmont, MA, 1995.

[3] R. Howard. *Dynamic Programming and Markov Processes*. MIT Press, Cambridge, MA, 1960.

[4] M. L. Littman. *Algorithms for Sequential Decision Making*. PhD thesis, Brown University, 1996.

[5] M. Melekopoglou and A. Condon. On the complexity of policy iteration for stochastic games. Technical Report CS-TR-90-941, Computer Sciences Department, University of Wisconsin, Madison, 1990.

[6] S. J. Russell and P. Norvig. *Artificial Intelligence: A Modern Approach*. Prentice Hall, Englewood Cliffs, New Jersey, 1995.

[7] R. S. Sutton and A. G. Barto. *Reinforcement Learning: An Introduction*. MIT Press, Cambridge, MA, 1998.